\documentclass[letterpaper, 10 pt, conference]{ieeeconf}  

\IEEEoverridecommandlockouts                              

\usepackage{hyperref}
\usepackage{calc}

\hypersetup{colorlinks,urlcolor=lightblue,citecolor={lightblue}}
\usepackage{wrapfig}
\usepackage{mathptmx} 
\usepackage[T1]{fontenc}
\usepackage{amsmath} 
\usepackage{multicol}
\usepackage{graphicx}
\usepackage{amssymb}
\usepackage{booktabs}
\usepackage{multirow}
\usepackage{subcaption}
\usepackage[export]{adjustbox}
\usepackage{wrapfig,lipsum,booktabs}
\let\labelindent\relax
\usepackage{enumitem}
\usepackage[backend=bibtex,natbib=true,sorting=none,giveninits=true,maxbibnames=99,url=false,doi=false]{biblatex}
\usepackage{color}
\usepackage[table,xcdraw]{xcolor}
\definecolor{turquoise}{cmyk}{0.65,0,0.1,0.3}
\definecolor{purple}{rgb}{0.65,0,0.65}
\definecolor{dark_green}{rgb}{0, 0.5, 0}
\definecolor{orange}{rgb}{0.8, 0.6, 0.2}
\definecolor{red}{rgb}{0.8, 0.2, 0.2}
\definecolor{darkred}{rgb}{0.6, 0.1, 0.05}
\definecolor{blueish}{rgb}{0.0, 0.3, .6}
\definecolor{light_gray}{rgb}{0.7, 0.7, .7}
\definecolor{pink}{rgb}{1, 0, 1}
\definecolor{greyblue}{rgb}{0.25, 0.25, 1}

\definecolor{lightblue}{RGB}{100, 170, 255} 
\definecolor{lightskyblue}{RGB}{220, 220, 250}
\newcommand{\ours}[0]{\rowcolor{lightskyblue}}

\usepackage{xcolor}

\usepackage[capitalize]{cleveref}
\crefname{section}{Sec.}{Secs.}
\Crefname{section}{Section}{Sections}
\Crefname{table}{Table}{Tables}
\crefname{table}{Table}{Table}
\crefname{figure}{Figure}{Figures}
\crefname{figure}{Fig.}{Figs.}
\Crefname{algocf}{Algorithm}{Algorithms}

\usepackage[noend]{algorithm2e}
\usepackage{algcompatible}%

\title{\LARGE \bf
Learning Sidewalk Autopilot from Multi-Scale Imitation with \\ Corrective Behavior Expansion
}

\author{
Honglin He$^{1}$,  Yukai Ma$^{1}$, Brad Squicciarini$^{2}$, Wayne Wu$^{1}$, Bolei Zhou$^{1}$ 
\thanks{$^{1}$Department of Computer Science, University of California, Los Angeles}%
\thanks{$^{2}$Coco Robotics} \\
\url{https://vail-ucla.github.io/MIMIC}
}

\begin{document}

\maketitle
\thispagestyle{plain}
\pagestyle{plain}





\begin{abstract}
Sidewalk micromobility is a promising solution for last-mile transportation, but current learning-based control methods struggle in complex urban environments. Imitation learning (IL) learns policies from human demonstrations, yet its reliance on fixed offline data often leads to compounding errors, limited robustness, and poor generalization. To address these challenges, we propose a framework that advances IL through corrective behavior expansion and multi-scale imitation learning. On the data side, we augment teleoperation datasets with diverse corrective behaviors and sensor augmentations to enable the policy to learn to recover from its own mistakes. On the model side, we introduce a multi-scale IL architecture that captures both short-horizon interactive behaviors and long-horizon goal-directed intentions via horizon-based trajectory clustering and hierarchical supervision. Real-world experiments show that our approach significantly improves robustness and generalization in diverse sidewalk scenarios. Demo video and additional information are available on the project page. 
\end{abstract}

\section{INTRODUCTION}
\label{sec:intro}

\begin{figure}[!t]
\centering
\vspace{-1em}
\includegraphics[width=\linewidth]{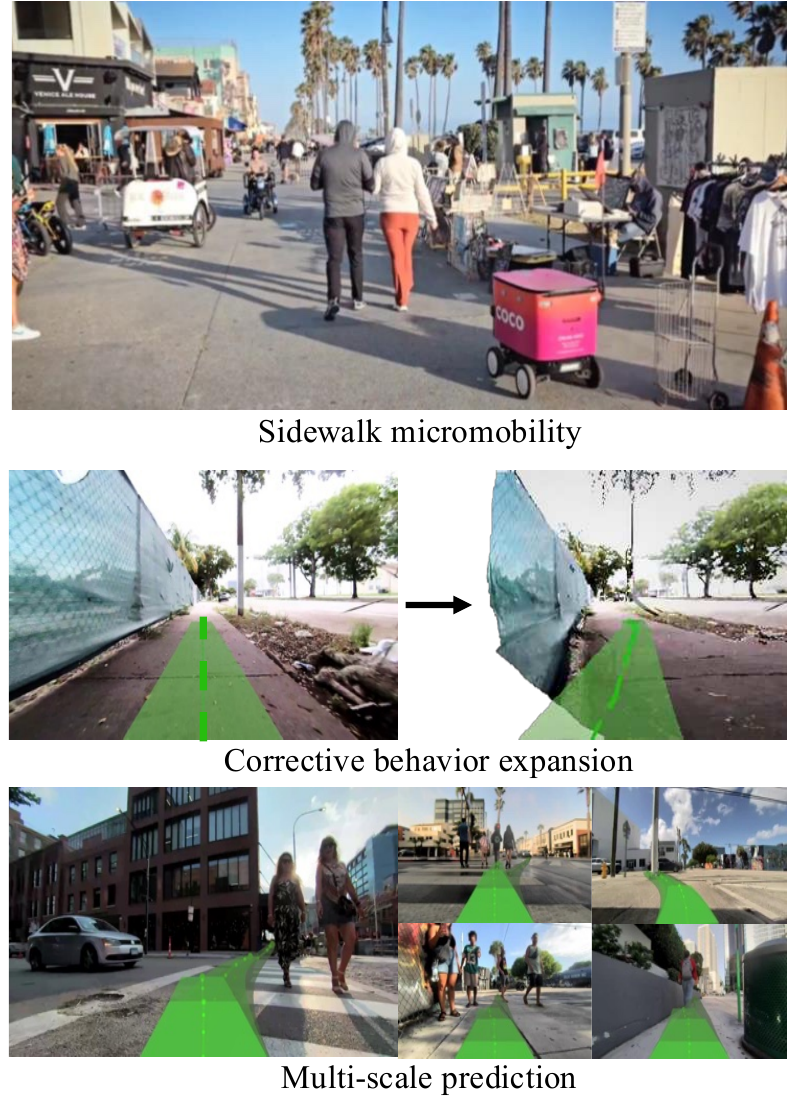}
\caption{
This work aims to utilize corrective behavior expansion and multi-scale prediction to learn an autopilot model for sidewalk micromobility.
}
\vspace{-2em}
\label{fig:teaser}
\end{figure}

Sidewalk micromobility has gained increasing attention as a solution for last-mile transportation in urban environments. Many applications have emerged in recent years, from robotic food delivery~\cite{engesser2023autonomous} to assistive power wheelchair~\cite{tuomi2021applications, liu2025service}. Figure~\ref{fig:teaser} shows a food delivery robot navigating a crowded sidewalk with pedestrians, street vendors, and other obstacles. With the rapid development of learning-based approaches, control and decision-making in these robot systems have moved beyond purely rule-based methods and increasingly relied on data-driven paradigms. A promising approach to sidewalk navigation is imitation learning (IL)~\cite{pomerleau1988alvinn}, which learns an end-to-end control policy directly from real-world human demonstrations. 
However, IL faces obvious limitations. Most notably, IL relies solely on learning from fixed and offline expert demonstrations, thus it often fails under closed-loop deployment where small errors are compounded over time and eventually lead to failure~\cite{lambert2022investigating}. Meanwhile, collecting demonstration data for deviated scenarios and critical corner cases is particularly difficult, further limiting the policy's robustness and generalizability. Beyond these, a practical challenge in sidewalk scenarios is that input observations are egocentric RGB videos, from which all information, including scene geometry and object semantics, must be inferred. It increases the difficulty of training a generalist sidewalk autopilot. In summary, policies trained purely with IL often work poorly in complex sidewalk environments.

Prior work has focused on scaling data volume~\cite{shah2022gnm,shah2023vint,sridhar2024nomad,liu2025citywalker, hirose2025learning} to address these challenges. However, much of the existing data has been collected in relatively simple or structured environments~\cite{shah2022gnm,shah2023vint}, which lack the complexity and diversity of real-world sidewalk scenarios. Meanwhile, these approaches are costly and still struggle to capture long-tail cases in specific domains, limiting generalization and robustness in real-world deployments. Other works utilize reinforcement learning (RL)~\cite{he2025seeing} to go beyond demonstrations. However, RL requires costly reward engineering and high-fidelity simulators, and often produces non-human-like behaviors. An alternative path has emerged from recent work~\cite{bansal2018chauffeurnet, goff2025learning}, where the data seen by the policy during IL can be extended by augmenting the training data distribution. This motivates our work: can we push IL further by generating diverse and plausible behaviors from a fixed offline dataset and fully exploiting each demonstration trajectory?



In this work, we study new ways to fully utilize real-world teleoperation data from both the \textit{data side} and the \textit{model side}, using data expansion with corrective behavior and multi-scale imitation learning. On the data side, we design a more effective way of augmenting teleoperation videos and demonstrations with corrective behaviors. Specifically, we synthesize novel data either from the observation side or by perturbing the action–observation–action loop, thereby exposing the policy to a broader distribution of plausible and diverse correction scenarios. Thus, the policy being trained can learn to recover from drifting off course. As illustrated in Fig.~\ref{fig:teaser}, our approach generates novel trajectories while preserving the underlying physical constraints in the original scenario. On the model side, we propose a multi-scale imitation learning and prediction framework to improve the policy's capacity to generalize across temporally and semantically diverse driving patterns. This framework first clusters trajectories based on temporal horizons and behavior patterns and then applies layer-wise supervision at different horizon levels, enabling the policy to learn both low-level interactions and high-level intentions in a unified framework. We summarize our contributions as: 
\begin{itemize}
    \item We propose a corrective behavior data expansion pipeline that synthesizes novel training data from existing teleoperation datasets by perturbing the action–observation–action loop, effectively increasing the coverage and diversity of training data.
    \item We propose a novel model architecture designed for tasks that require both short-horizon interactive behaviors and long-horizon goal-directed intentions.
    \item We establish real-world deployment and validation, demonstrating that our approach improves policy robustness and generalization in diverse, complex sidewalk environments using only offline teleoperation data.
\end{itemize}

\section{Related Work}
\label{sec:related work}


\noindent\textbf{Sidewalk navigation.} Visual navigation has a long history. Early works focused on leveraging constructed 3D maps for localization and planning~\cite{kummerle2011g,lei2025gaussnav}. In contrast, recent advances increasingly favor end-to-end learning models that map raw sensory observations directly to actions~\cite{shah2022gnm,shah2023vint,sridhar2024nomad,liu2025citywalker,hirose2025learning}, known as mapless navigation. While these approaches span a wide range of navigation tasks, sidewalk navigation presents unique challenges, including narrow passages, frequent dynamic interactions with diverse pedestrians and other moving objects such as scooters and bikers, complex structures such as curbs and crosswalks, and complex urban layouts. Given these challenges, traditional map-based approaches, which rely on offline map construction, are often brittle in such environments. In this work, we focus on data-driven urban navigation foundation models that generalize across diverse sidewalk scenarios under varying environmental conditions. Prior work has collected large-scale data from real-world settings for policy learning~\cite {sridhar2024nomad, liu2025citywalker}. However, most of these approaches and datasets are limited to either indoor environments, outdoor but sparsely populated scenarios, or driving scenarios. While some prior studies have claimed that point-goal navigation is largely solved~\cite {puig2023habitat}, the inherent complexity of real-world sidewalk navigation in a mapless, monocular RGB-camera setting remains a significant challenge that this work aims to address. 


\noindent\textbf{Learning from teleoperation data.} Teleoperation provides a practical way to collect large-scale demonstrations for policy learning across diverse tasks and embodiments~\cite{sridhar2024nomad,liu2025citywalker}. Early efforts focused on modular learning, \textit{i.e.} training different models for each sub-task like learning object detectors~\cite{ren2015faster}, planners~\cite{qureshi2019motion} and controllers~\cite{mittal2023orbit} separately. Recently, increasing attention has been paid to end-to-end approaches~\cite{sridhar2024nomad, liu2025citywalker}. These end-to-end approaches eliminate the need for handcrafted modules and offer the potential to capture complex correlations within the data. These offline end-to-end learning approaches require large volumes of data for training. However, in some real-world scenarios, data cannot be effectively collected or fully exploited due to limitations like coverage or annotation quality.
At the same time, prior work has shown that imitation-only policies degrade rapidly when facing covariate shift or compounding errors~\cite{ross2011reduction, bansal2018chauffeurnet}. These challenges have led researchers to explore alternative strategies. 
In particular, many approaches have been developed to learn from a mixture of offline demonstrations and online interactions, combining the strengths of imitation learning and reinforcement learning to improve policy robustness and adaptability, including DAgger~\cite{ross2011reduction}, residual reinforcement learning~\cite{wang2025x}, and RLHF~\cite{peng2023learning, peng2025data}. Our work focuses on end-to-end learning without relying on reinforcement learning. Instead, we synthesize training data containing deviation-recovery trajectories, enabling the model to learn a robust policy that mitigates compounding errors commonly encountered in imitation learning. We also introduce a novel architecture tailored for tasks that require both short-horizon interactive behaviors and long-horizon goal-directed intentions, and demonstrate its effectiveness in learning from synthesized deviation-recovery trajectories.

\section{Method}
\label{sec:method}



In this section, we introduce the proposed learning framework \textbf{MIMIC} (\textbf{M}ulti-scale \textbf{IMI}tation with \textbf{C}orrective expansions), which leverages pretrained models to generate out-of-domain scenarios by training on both expert demonstrations and near-failure experiences, using multi-scale imitation.

\begin{figure*}[t!]
\centering
\includegraphics[width=\textwidth]{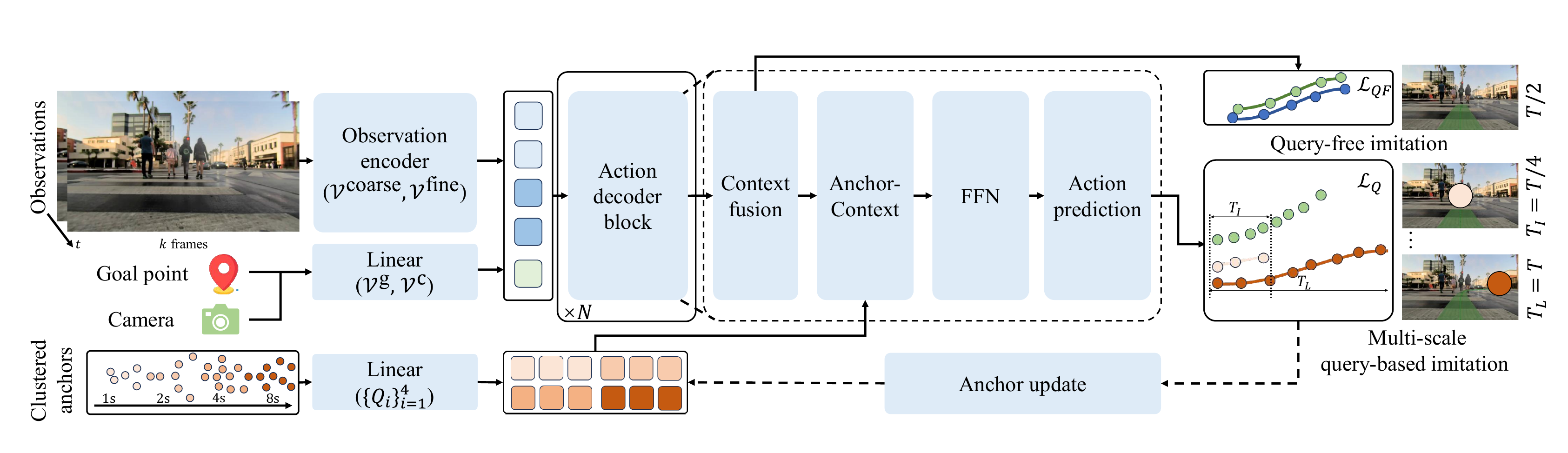}
\caption{\textbf{Illustration of the MIMIC framework}. The model adopts an encoder–decoder architecture that combines coarse historical embeddings with fine-grained current visual observations as context. The context encoder converts the observation sequence by combining the coarse flattened features of historical frames with the fine patch-level features of the current frame, together with the goal point and camera features. The action decoder leverages time-horizon-specific anchors to produce actions parameterized by GMMs across multiple horizons, thereby enhancing the output's diversity and robustness.
}
\label{fig:Framework}
\vspace{-5mm}
\end{figure*}

\subsection{Problem Formulation}

We aim to train a policy for mapless point-goal visual navigation, in which the agent receives only egocentric RGB images and GPS signals as input, both readily available on real-world robots. This setting eliminates the need for pre-built maps or localization modules and can be viewed as a sequential decision-making problem under partial observability.
At each timestep $t$, the agent is provided with a history of the past $T_h$ RGB observations $i_{t-T_h:t}$, its past $T_h$ ego-states $e_{t-T_h:t}$ (e.g., GPS locations, velocities, orientations), and a sub-goal or route $g_t$ expressed in ego-centric coordinates. The policy $\pi_{\theta}$ takes observation $o_t=(i_{t-T_h:t},e_{t-T_h:t},g_t)$ as input and gives the action $a_t$ to control the robot. In the paradigm of imitation learning, the goal is to train a policy $\pi_{\theta}$ by minimizing the discrepancy between the agent’s actions and the expert demonstrations. Formally, given expert trajectories $\mathcal{D} = {(o_{t}, a_t)}_{t=0}^T$, the objective is $\min_{\theta}\mathbb{E}_{(o_{t}, a_t)\sim\mathcal{D}}[\mathcal{L}(\pi_{\theta}(o_{t}),a_t)]$. 
In our formulation, $\pi_{\theta}$ outputs a probability distribution over candidate actions, and we adopt the negative log-likelihood (NLL) loss for supervision, \textit{i.e.},
\vspace{-2mm}
\begin{align}
    \mathcal{L}(\pi_{\theta}(o_{t}),a_t)=-\log\pi_{\theta}(a_t|o_{t}).
\end{align}
For the action space $\mathcal{A}$, we define it as a sequence of waypoints sampled at a fixed frame rate. Each action $a_t \in \mathcal{A}\subset \mathbb{R}^{T\times3}$ corresponds to a trajectory segment represented in bird’s-eye view (BEV), where each waypoint encodes a 2D location and an orientation in ego-centric coordinates. To parametrize the model for the action distribution, we use a Gaussian Mixture Model (GMM)~\cite{shi2022motion}. Specifically, at each timestep $t$, the policy $\pi_\theta$ outputs the parameters of a mixture distribution.
\vspace{-2mm}
\begin{align}
    \pi_{\theta}(a_t|o_{t})=\sum_{m=1}^{M}p_{\theta,m}(o_t)\mathcal{N}(\mu_{\theta,m}(o_t),\sigma_{\theta,m}(o_t)),
\end{align}
\noindent where $\mu_{\theta,m}(o_t)=\left\{(\hat{x}_{t+\tau},\hat{y}_{t+\tau},\hat{\psi}_{t+\tau})\right\}_{\tau=1}^{T}$ denotes the predicted waypoint sequence (2D position and heading) over horizon $T$ for the $m$-th Gaussian component conditioned on observation $o_t$ and $\sigma_{\theta,m}(o_t)$ denotes the corresponding variance capturing the uncertainty of the predicted waypoints.




\subsection{Multi-scale Imitation Learning with Anchors}


\noindent\textbf{Multi-scale supervision.} Before introducing the model architecture, we first present the key modeling of the action space in our framework. While many existing imitation learning methods supervise the policy via the difference between the ground truth and model outputs at a single temporal scale, typically focusing on short-term predictions to ensure immediate responsiveness. However, this paradigm often leads to shortcut learning~\cite{geirhos2020shortcut}, where the model relies on spurious correlations rather than learning the intended underlying meaningful representations. Such behavior is particularly problematic in navigation tasks, which require both fine-grained interaction and global consistency to handle complex urban environments with many pedestrians, vehicles, road structures, etc. Therefore, we argue for introducing a multi-scale action space from short-horizon to long-horizon, where the policy is explicitly supervised across multiple temporal scales, enabling it to learn both immediate behaviors and long-term goal-aligned behaviors within a unified framework.

Concretely, we enrich the action space $\mathcal{A}$ by incorporating a multi-level supervision across different temporal horizons. Instead of supervising the policy at a single scale, we provide guidance simultaneously at the immediate, short, medium, and long horizons, denoted as $\left\{\mathcal{A}_1, \mathcal{A}_2, \mathcal{A}_3, \mathcal{A}_4\right\}$. This hierarchical supervision mitigates the shortcut behavior observed with single-horizon training~\cite{liu2025citywalker} — where the model tends to optimize only for immediate success. As a result, the policy is encouraged to align fine-grained reactivity with long-term planning, yielding a more expressive and stable navigation model. Specifically, $a_{t,i}=\left\{(x_{t+\tau},y_{t+\tau},\psi_{t+\tau})\right\}_{\tau=1}^{T_i}\in\mathcal{A}_i$ and $\left\{T_1=\frac{T}{8}, T_2=\frac{T}{4}, T_3=\frac{T}{2}, T_4=T\right\}$ in our setting.

\noindent\textbf{Model architecture.} As shown in Fig.~\ref{fig:Framework},
we adopt an encoder–decoder architecture to model the policy. The encoder processes multimodal inputs—RGB observations, ego-states, and goal signals—into a compact spatiotemporal representation. Specifically, we encode the history of image observations $i_{t-k:t-1}$ using a visual backbone initialized from DINOv3~\cite{simeoni2025dinov3}. Each historical image within the $T_h$ input frames is first encoded into a high-dimensional embedding, forming a coarse temporal feature sequence $\mathcal{V}^{\text{coarse}} \in \mathbb{R}^{T_h \times C}$. For the current image observation $i_t$, we extract patch-level features from the backbone initialized from DINOv3~\cite{simeoni2025dinov3}, and image patches are then downsampled via grid pooling and flattened into a sequence of tokens $\mathcal{V}^{\text{fine}} \in \mathbb{R}^{64 \times C}$, which preserve fine-grained spatial details such as obstacles and scene geometry. The navigation goal is modeled as a compact 3D vector $(d,\cos\phi_{\text{goal}},\sin\phi_{\text{goal}})$, encoding the distance and relative orientation to the target. Camera intrinsic parameters, together with the camera's 3D location relative to the robot center, are denoted by $c\in\mathbb{R}^{16}$. Both goal and camera parameters are projected into the embedding space $\mathcal{V}^{g},\mathcal{V}^{c}\in\mathbb{R}^{1\times C}$ using an MLP. Each coarse visual token $\mathcal{V}^{coarse}_i$ is first modulated via a FiLM layer~\cite{perez2018film} to incorporate conditioning temporal information $\mathcal{V}^{coarse}_i\leftarrow\mathcal{V}^{coarse}_i\odot\gamma_i+\beta_i$, where $(\gamma_{i}, \beta_{i})\in\mathbb{R}^{C}$ are scaling and shifting parameters generated from time-step $t-i$ relative to the current frame. 


The action decoder comprises a stack of context-fusion and trajectory-refinement layers. At each layer, we first fuse the context features $\mathcal{V}=[\mathcal{V}^{coarse},\mathcal{V}^{fine}, \mathcal{V}^{g},\mathcal{V}^{c}]$ via multi-head attention~\cite{vaswani2017attention} $\mathcal{V}'= \text{MHA}(Q=\mathcal{V};K,V=\mathcal{V})$. Subsequently, the context features are used as keys and values for action decoding, allowing the decoder to attend to relevant spatial-temporal cues during trajectory prediction. The decoder generates actions by referencing a set of anchor trajectories, which serve as structured priors for plausible motion patterns. These anchors are pre-generated from data statistics based on K-means~\cite{lloyd1982least}. More specifically, instead of relying on a single query set, we generate four scale-specific anchor sets $\left\{\mathbb{A}_i\right\}_{i=1}^4;\ \mathbb{A}_i\subset\mathbb{R}^{64\times3},\ \forall i$ that correspond to the immediate, short, medium, and long horizons. They would be mapped to query tokens $\left\{\mathcal{Q}_i\right\}_{i=1}^4$ via a linear layer. Each query set interacts with the encoder representation $\mathcal{Q}_i'=\text{MHA}(Q=\mathcal{Q}_i;K,V=\mathcal{V}')$, enabling the model to jointly capture local reactivity and global consistency across different temporal scales. Given the multi-scale queries $\left\{\mathcal{Q}_i\right\}_{i=1}^4$ and the context condition $\mathcal{V}$ as input, $k$-th decoding layer $\mathcal{F}_{k,\theta}$ generates five trajectory predictions: four query-based heads, each conditioned on a specific query and the context, and one query-free head that relies solely on the contextual information $\mathcal{V}'$, \textit{i.e.},

\vspace{-5mm}
\begin{align}
\left\{\hat{\mathcal{T}}_{\text{QF}},\hat{\mathcal{T}}_{\text{Q}}\right\}&= \mathcal{F}_{k,\theta}(\mathcal{Q}_{1:4}, \mathcal{V}), \\
\hat{\mathcal{T}}_{\text{Q}}&=\left\{\hat{p}_{i,m}, \hat{\mathcal{T}}_{i,m}\right\}_{i=1:4,\,m=1:M},
\end{align}
\vspace{-5mm}

\noindent where $\hat{\mathcal{T}}{i,m}$ denotes the predicted trajectories and $\hat{p}{i,m}$ the corresponding confidence scores of mode $m$ at horizon $i$, and $\hat{\mathcal{T}}_{QF}$ is the query-free trajectory prediction.

For each data sample, we assign a positive label to the mode $h_i$ within the candidate trajectory set $\left\{\hat{\mathcal{T}}{i,m}\right\}_{m=1}^{M}$ at horizon $i$, where the selected anchor trajectory $\hat{\mathcal{T}}{i,h_i}$ has the closest end-point to the ground-truth trajectory $\mathcal{T}{i}^{\text{gt}}$. That is, $p_{i,h_i} = 1$ and $p_{i,m} = 0$ for all $m \ne h_i$.
For simplicity, we assume a fixed covariance $\Sigma = 0$, such that each trajectory mode degenerates into a deterministic prediction. In parallel, we introduce an auxiliary query-free (QF) reconstruction task that directly predicts future actions from the encoded visual patches, without relying on decoder queries. The QF head generates a single trajectory at a fixed short-term horizon (e.g., $\frac{T}{4}$), promoting fine-grained short-horizon supervision.

\vspace{-5mm}
\begin{align}
    \mathcal{L}_{k} &=\mathcal{L}_{k,Q}+\mathcal{L}_{k,QF}, \\
    \mathcal{L}_{k,Q} &= \sum_{i=1}^4\sum_{m=1}^{M}[\mathcal{L}_{k,i;reg}+\lambda\cdot\mathcal{L}_{k,i;cls}],
\end{align}

\noindent where $\mathcal{L}_{k,QF}$ and $\mathcal{L}_{k,i;reg}$ are regression loss terms between the prediction and ground truth, and $\mathcal{L}_{k,i;cls}$ is the BCE loss between $p_{i,h_i}$ and $\hat{p}_{i,h_i}$. Finally, the overall training objective averages the supervision over all $K$ decoder layers.

\subsection{Teleoperation Data Expansions}

\begin{figure}[!t]
\centering
\includegraphics[width=\linewidth]{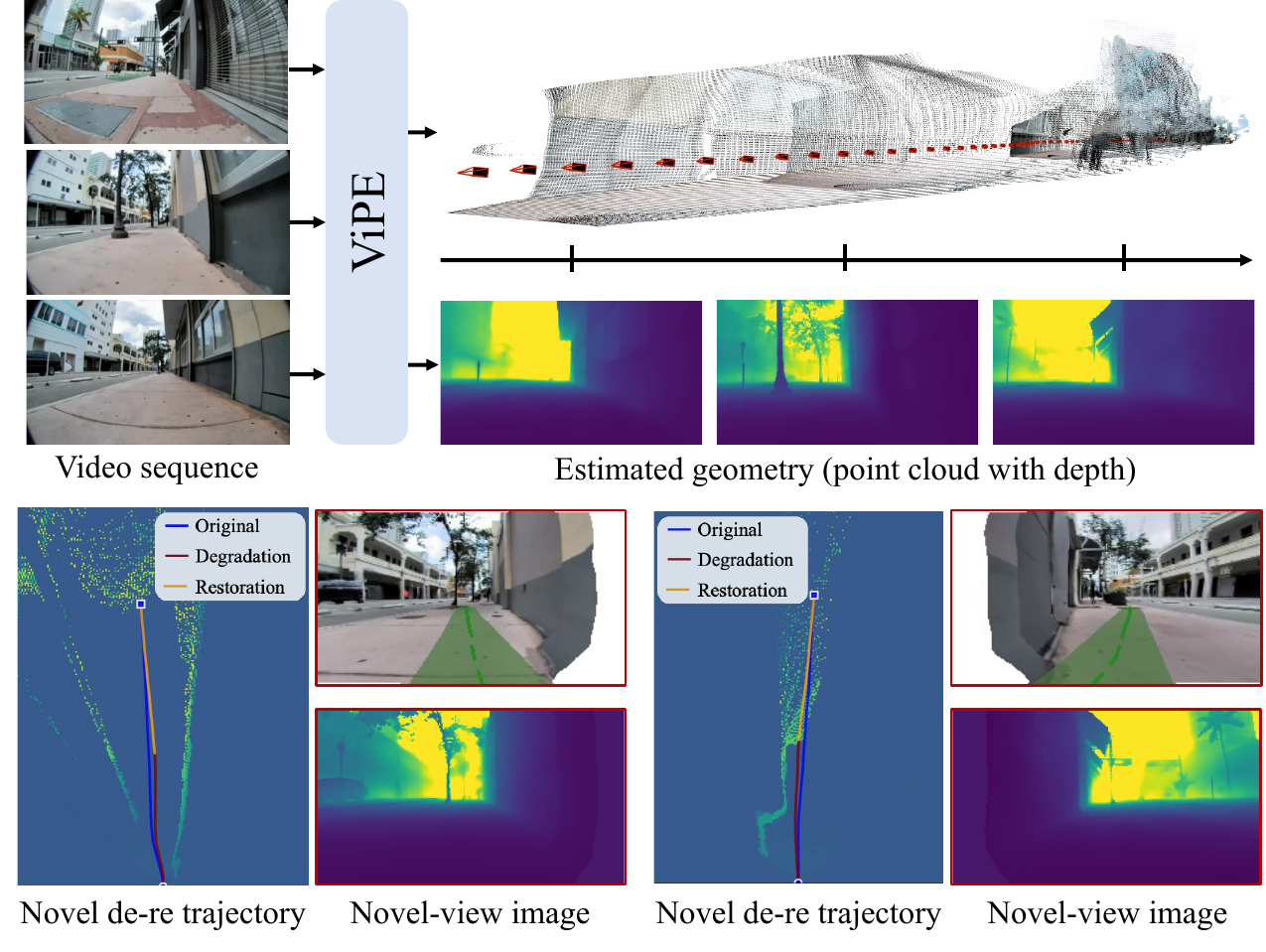}
\caption{ 
Illustration of the corrective behavior expansion. We first estimate the depth sequence and reconstruct a point cloud. Given the 3D point cloud, we perturb the trajectory using a deviation–recovery noise sequence. Then we synthesize corresponding observation-action pairs.
}

\label{fig:behavior-per}
\vspace{-5mm}
\end{figure}

\noindent\textbf{Corrective behavior expansions.}  
Since the recorded logs are dominated by normal and straightforward observations and actions, they rarely include demonstrations that show how to recover from failure or near-failure cases~\cite{bansal2018chauffeurnet,goff2025learning}, for instance, the corrective actions to take when a vehicle starts drifting off its intended path. As a result, a policy trained purely by imitating demonstration data cannot learn to recover from its own mistakes. To simulate such failure-correction scenarios, we deliberately generate trajectories in which the model would take incorrect actions (e.g., deviating from the intended route, stepping onto the grass, colliding with obstacles, or stopping prematurely), and then provide corrective actions as supervision. As illustrated in Fig.~\ref{fig:behavior-per}, we begin by estimating a continuous metric depth sequence $I_D\in\mathbb{R}^{(T_h+T)\times H\times W}$ from ViPE~\cite{huang2025vipe} to annotate the surrounding scene geometry. After that, we leverage depth and RGB observations to construct a colored point cloud sequence $\mathcal{P}\in\mathbb{R}^{(T_h+T)\times (H\times W) \times 6}$ in the ego-centric frame, providing a 3D geometric representation of the scene, which we use to perturb trajectories. To induce deviations, we define a shifting sequence $\Delta\mathcal{T}(\tau)$ that smoothly varies from $0$ back to $0$ over the prediction horizon, following a sine-like profile $\Delta\mathcal{T}(\tau)=\alpha\cdot\text{sin}(\frac{\pi\cdot \tau}{T_h+T})$, where $\alpha$ controls the maximum displacement. The novel RGB observations $i'_{t-T_h:t}$ are synthesized under the perturbation $\left\{\Delta\mathcal{T}(\tau)|\tau=t-T_h, ..., t-1\right\}$ by reprojecting the colored point cloud sequence $\mathcal{P}$ into the ego-centric camera frame, conditioned on the perturbed trajectories. Given the perturbed observation sequence, the supervision trajectory is the recovery trajectory generated from the original one and shifted by $\left\{\Delta\mathcal{T}(\tau)|\tau=t,...,t+T-1\right\}$. This perturbation scheme introduces temporary lateral or longitudinal drifts into the original expert trajectory, mimicking realistic failure cases such as veering off-road or hesitating at obstacles. By pairing each perturbed trajectory with a corrective failure-to-recovery maneuver, we obtain failure–correction pairs that enable the policy to learn robust recovery behaviors.

\noindent\textbf{Sensor augmentation.} Besides the lack of corrective behaviors in the collected teleoperation dataset, the visual appearance of recorded videos is often overly simple, with fixed lighting, limited weather conditions, and low diversity of backgrounds. More importantly, teleoperation logs from the real world often over-represent normal behaviors (e.g., straight-line movement on clear sidewalks) while under-representing rare but safety-critical events, such as erroneous operations where the robot steps onto the grass, or pauses at crowded intersections. To address data imbalance, we introduce generative augmentation to enrich both the sensory inputs and the state–action pairs. 

\begin{figure}[t!]
\centering
\includegraphics[width=\linewidth]{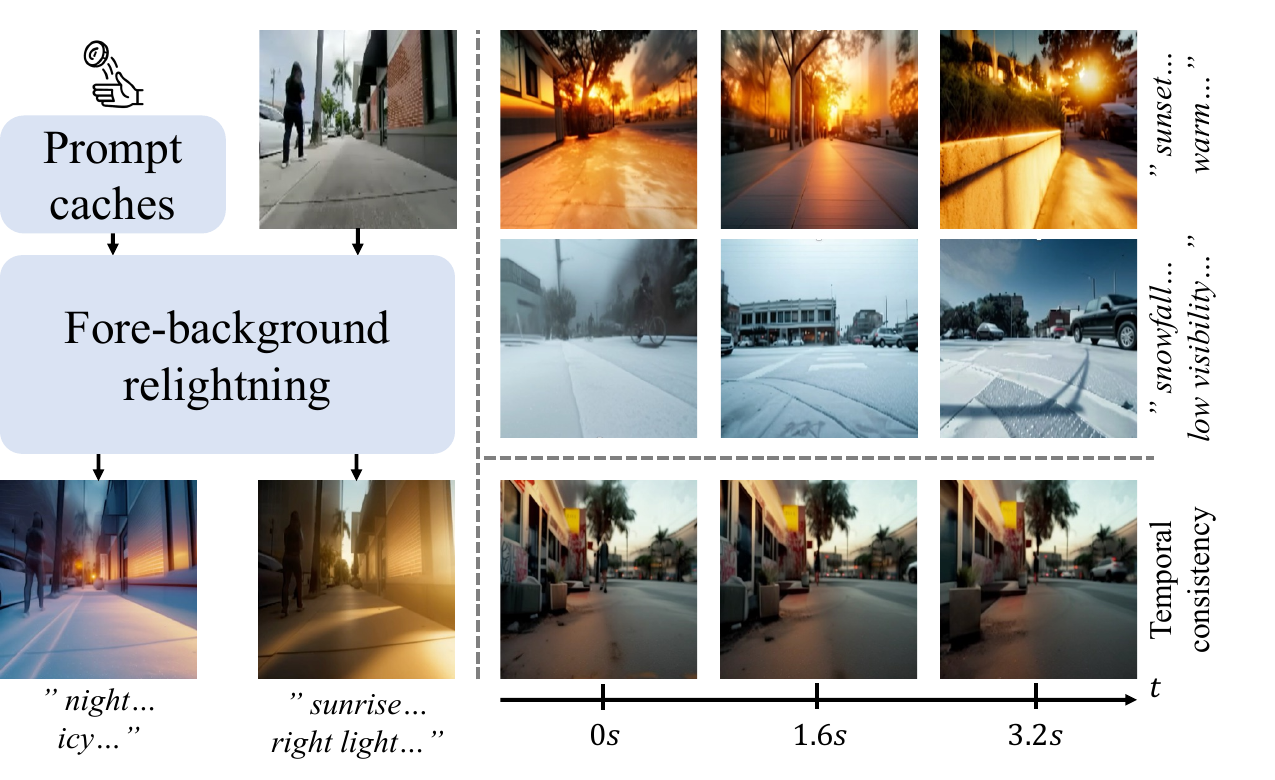}
\caption{
Illustration of the sensor augmentation. A pretrained relighting model is used to modify the scene guided by different lighting prompts. The original scenario is segmented into foreground and background regions where different relighting parameters are applied. The outputs are then blended to synthesize novel relighted observations.
}
\vspace{-5mm}
\label{fig:appearance-per}
\end{figure}

 The key principle is to preserve scene geometry and structure while altering visual appearance. Prior work commonly employs depth- or semantic-based re-rendering~\cite{agarwal2025cosmos,alhaija2025cosmos} to diversify illumination and textures. However, these approaches often introduce artifacts, such as inconsistent blending, where nearby objects inherit background lighting conditions. To alleviate this issue, we adopt a relighting model, Light-A-Video~\cite{zhou2025light},  that preserves scene geometry while modifying global appearance. Specifically, the model disentangles foreground objects $I_{f}$ from the background $I_b$ using depth, applies prompt-based relighting with different strength coefficients to the foreground and background, \textit{i.e.}, 

\vspace{-5mm}
\begin{align}
     I' &= f_{\text{relight}}(I_{f};\,\alpha_{f},\,p) \oplus f_{\text{relight}}(I_{b};\,\alpha_{b},\,p),\ \alpha_f < \alpha_b,
\end{align}

\noindent where $f_{\text{relight}}(\cdot)$ denotes the prompt-based relighting model, 
$p$ is the textual prompt controlling illumination style, 
and $\alpha_f=0.1, \alpha_b=0.5$ are the respective relighting strengths applied to the foreground and background. 
As shown in Fig.~\ref{fig:appearance-per}, this asymmetric design preserves foreground consistency while enhancing background diversity.

\vspace{-1mm}
\section{Experiments}
\label{sec:experiments}

eWe evaluate our proposed approach, MIMIC, on both offline sidewalk videos and real-world deployments with a wheeled robot. We report the overall performance of our model in comparison with prior baselines, conduct ablation studies to analyze the contributions of all components, and provide qualitative results to illustrate the effectiveness of the proposed approach.

\subsection{Dataset} 

\begin{figure}[t!]
\centering
\includegraphics[width=\linewidth]{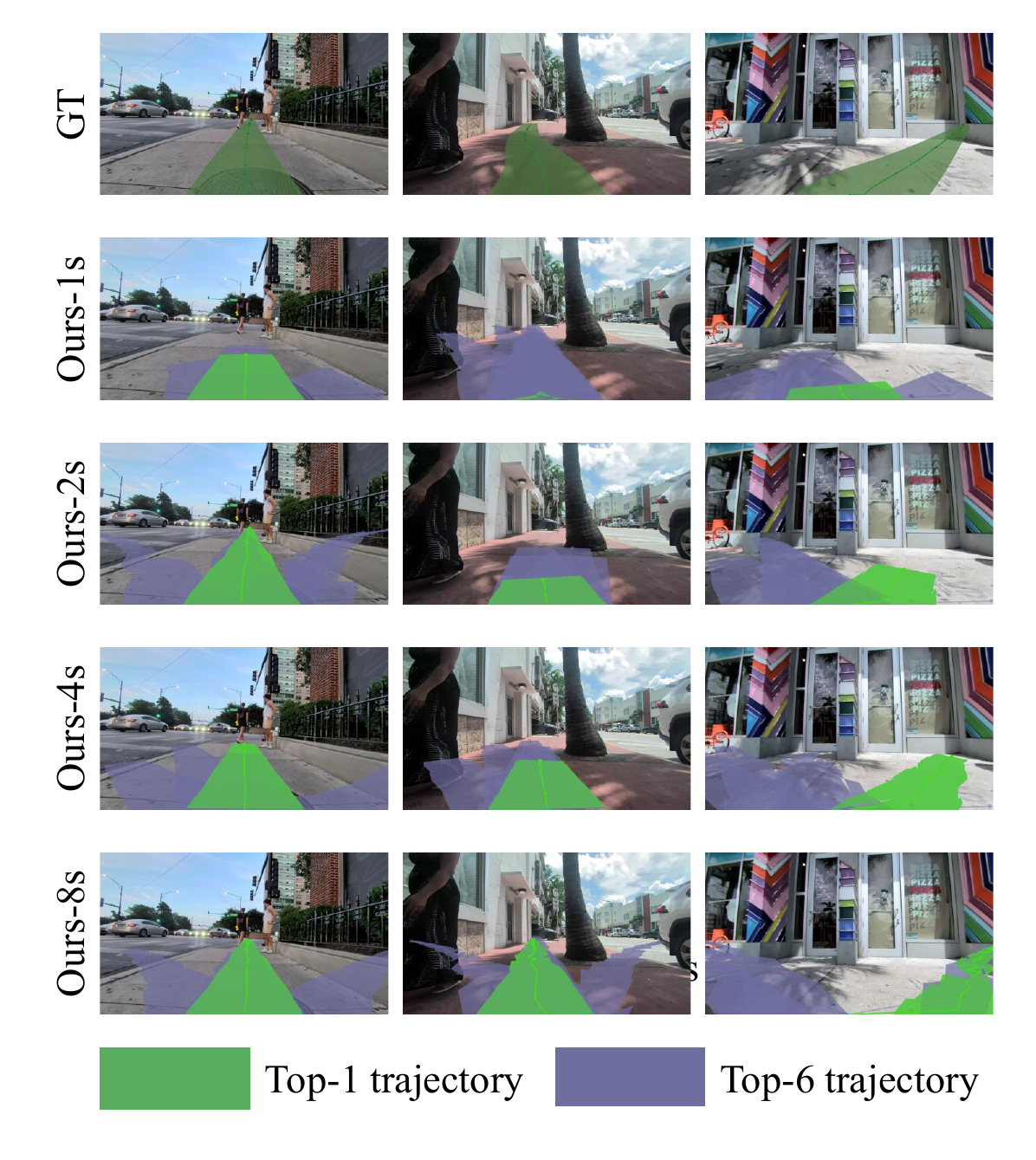}
\caption{
Qualitative results of MIMIC on the \textbf{CoS} test set. The green trajectory denotes the one with the highest probability, while the others represent the top-6 trajectories filtered by non-maximum suppression (NMS).
}
\vspace{-5mm}
\label{fig:qualitative}
\end{figure}

We have collected a large-scale video teleoperation dataset, \textbf{CoS} (short for Coco-on-SideWalks).
In total, the dataset contains 3,040 trajectories collected by multiple wheeled robots from Coco Robotics\footnote{\url{https://www.cocodelivery.com/}} navigating diverse sidewalks across various US cities, each lasting 1 minute, amounting to about 50 hours of data. For each trajectory segment, we record fisheye RGB videos at 20Hz, along with synchronized robot-state logs that include position, orientation, linear velocity, and angular velocity, derived from GPS and onboard odometry. We split the dataset into 2,740 trajectories for training, 200 for validation, and 100 for testing. As illustrated in Fig.~\ref{fig:qualitative}, we present qualitative results of predicted trajectories alongside ground truth across several scenarios in the dataset. 

\noindent\textbf{Dataset curation.} After collecting teleoperation logs, we perform a systematic curation process to ensure data quality and consistency. Specifically, the process involves:

\noindent (i) Behavior classification and balancing.
We classify trajectories into basic behavioral categories (e.g., straight walking, turning, stopping). Since straightforward walking behaviors dominate the teleoperation logs, we downsample redundant segments while retaining a higher proportion of diverse behaviors, thereby alleviating class imbalance.

\noindent (ii) Filtering abnormal segments.
We remove sequences in which the robot exhibits undesirable motions, such as sensor-induced rotations while staying still or backward behaviors. This filtering step prevents the model from overfitting to noisy or unrepresentative actions.

\noindent (iii) Goal point definition.
For each trajectory, the goal point is defined in two ways: (1) randomly sampling the next 5–20 frames like~\cite{sridhar2024nomad,liu2025citywalker}, or (2) splitting the trajectory into $N$ segments ($N\in[3,7]$) and selecting the nearest segment endpoint. This strategy avoids shortcut learning by sampling not only the immediate few frames that are strongly correlated with the current state.

\noindent (iv) Trajectory smoothing.
For each sub-trajectory of length $(T_h+T)$ used in training, we apply slerp to smooth the recorded poses, thereby reducing variations caused by differences among teleoperators and noise introduced by operation habits. Specifically, we first compute the total trajectory length and then regenerate the trajectory by interpolating poses at a constant velocity along the path.

\subsection{Implementation Details} 

Our neural network consists of 4 encoder–decoder layers with a hidden dimension of 512. The observation encoder is initialized from DinoV3-S~\cite{simeoni2025dinov3}. Each input sequence consists of 16 frames sampled at 5 Hz, with all images resized to a resolution of $256 \times 256$. For trajectory prediction, we define the longest horizon as 40 frames at 5Hz, and each horizon is associated with 64 anchors for multi-modal decoding. All parameters are trained jointly in an end-to-end manner. 

We adopt a cosine learning rate schedule with an initial learning rate of $1\times 10^{-4}$ and a total batch size of 192. To improve the model robustness, we apply random masking during training: the goal token is masked with a probability of 0.5 to force the model to exploit contextual features, while other tokens are masked with a probability of 0.2.   The model is trained for 100 epochs, which takes approximately 1.5 days on 8 NVIDIA L40S GPUs. 






\begin{table}[!t]
\centering
\caption{{Open-loop evaluation on \textbf{CoS-Regular}.} 
}
\vspace{-0.5em}
\label{tab:openloop}
\resizebox{\linewidth}{!}{%
\begin{tabular}{l ccccc}
\toprule
 & minADE$_{1s}$~$\downarrow$ & minFDE$_{1s}$~$\downarrow$  & mAP~$\uparrow$ & L2$_{1s}$~$\downarrow$ &  L2$_{2s}$~$\downarrow$ \\
\midrule
GNM$^\ddagger$ &0.594 &0.988 &- &0.988 &-\\
ViNT$^\ddagger$ &0.638 &1.056 &- &1.056 &-\\
NoMaD$^\ddagger$ &0.523 &0.858 &0.216 &1.072 &2.182\\
\midrule
MBRA &0.617 &1.019 &- &1.019 &2.034\\
CityWalker &0.648 &1.125 &-&1.125 &-\\

\midrule
ViNT* &0.247 &0.425 &- &0.425 &0.925\\
CityWalker* &0.180 &0.353 &-&0.353 &0.786\\
\midrule
\ours MIMIC &\textbf{0.071} &\textbf{0.129} &\textbf{0.695} &\textbf{0.342} &\textbf{0.700}\\ 
\bottomrule

\end{tabular}%

}
\end{table}




\begin{table}[!t]
\centering
\caption{{Open-loop evaluation on \textbf{CoS-Recovery}.} }
\vspace{-0.5em}
\label{tab:openloop2}
\resizebox{\linewidth}{!}{%
\begin{tabular}{l ccccc}
\toprule
 & minADE$_{1s}$~$\downarrow$ & minFDE$_{1s}$~$\downarrow$  & mAP~$\uparrow$ & L2$_{1s}$~$\downarrow$ &  L2$_{2s}$~$\downarrow$ \\
\midrule

MBRA &2.586 &3.297 &- &3.297 &7.000\\
CityWalker  &0.929 &1.740 &-&1.740
&-\\
CityWalker* &0.398 &0.695 &- &0.695
&1.397\\
\midrule
\ours MIMIC &\textbf{0.196} &\textbf{0.328} &\textbf{0.565} & \textbf{0.645} & \textbf{1.348}\\ 
\bottomrule

\end{tabular}%
\vspace{-5mm}
}
\end{table}

\subsection{Open-Loop Evaluation}

We first evaluate our approach in an open-loop setting, where predicted trajectories are compared against ground-truth future trajectories on the test set. We conduct experiments on two subsets of our dataset: \textbf{CoS-Regular} and \textbf{CoS-Recovery}. The SideWalks-Regular set contains normal teleoperation trajectories, while the SideWalks-Recovery set includes perturbed observations. This separation allows us to evaluate both the prediction accuracy under standard conditions and the robustness of the policy when confronted with deviation-induced observations. For evaluation, we adopt the standard open-loop metrics proposed in prior works~\cite{varadarajan2022multipath++, ettinger2021large}. A trajectory is considered positive if its endpoint at 1s lies within 1m of the ground truth. It is worth noting that previous works generate only a single-mode trajectory. Therefore, when reporting mAP, we report \underline{them} using only the Average Precision (AP).

\noindent\textbf{Baselines.}We compare against several state-of-the-art navigation foundation models: 1) image-goal approaches$^\ddagger$ including GNM~\cite{shah2022gnm}, ViNT~\cite{shah2023vint}, NoMaD~\cite{sridhar2024nomad}, and 2) point-based approaches CityWalker~\cite{liu2025citywalker}, MBRA~\cite{hirose2025learning}, ViNT* and CityWalker* (*denotes model re-trained on our dataset). 

%
Tab.~\ref{tab:openloop} and Tab.~\ref{tab:openloop2} show that MIMIC consistently outperforms all baseline methods on both Regular and Recovery test sets. Specifically, MIMIC achieves a 60.6\% lower minADE\textsubscript{1s} and 63.5\% lower minFDE\textsubscript{1s} than the second-best method (CityWalker*) on SideWalks-Regular, along with a 19.5\% improvement in L2\textsubscript{2s}. On the SideWalks-Recovery set, MIMIC yields a 50.8\% reduction in minADE\textsubscript{1s} and 52.8\% in minFDE\textsubscript{1s} compared to CityWalker*, while also achieving a 3.5\% lower L2\textsubscript{2s}. 

We provide qualitative results of our approach on \textit{Sidewalks}. As illustrated in Fig.~\ref{fig:qualitative}, the predictions remain accurate across all horizons. In the second column, our approach successfully finds a feasible path between the pedestrian and the obstacle. In the third column, when encountering a door in front, the policy attempts to avoid a collision.
 
\subsection{Ablation Study}

We conduct ablation studies to evaluate the effectiveness of the model design and the data expansions.






\begin{table}[!t]
\centering
\caption{Ablation study on model design. $I,S,M,L$ denote the prediction head at immediate, short, medium, and long horizons, respectively, and $QF$ denotes the prediction head derived directly from the context features.
}
\vspace{-0.5em}
\label{tab:model abl}
\resizebox{\linewidth}{!}{%
\begin{tabular}{l l l l l ccccc}
\toprule
  $I$& $S$ &$M$ &$L$ &$QF$ & minADE$_{1s}$~$\downarrow$ & minFDE$_{1s}$~$\downarrow$  & mAP~$\uparrow$ & L2$_{2s}$~$\downarrow$ &  L2$_{8s}$~$\downarrow$ \\
\midrule
& & & &$\checkmark$ &0.188 &0.371 &- &0.789 &4.083\\
&$\checkmark$ & & & &\textbf{0.067} &\textbf{0.113} &0.310 &\textbf{0.596} &-\\ 
& & &$\checkmark$ & &0.081 &0.132 &0.670 &0.711 &3.805\\ 
$\checkmark$ &$\checkmark$ &$\checkmark$ &$\checkmark$ & &0.074 &0.135 &0.680 &{0.637} &4.068\\ 
$\checkmark$ &$\checkmark$ &$\checkmark$ &$\checkmark$ &$\checkmark$ &{0.071} &{0.129} &\textbf{0.695} &0.700 &\textbf{3.718}\\ 
\bottomrule
\end{tabular}%
}
\end{table}

\noindent\textbf{Effect of the model design.} We conduct ablation studies by comparing different model configurations. As illustrated in Tab.~\ref{tab:model abl}, introducing anchor-based prediction $S$ significantly improves short-term accuracy compared to relying solely on the context-based head ($QF$). The short-horizon head ($S$) achieves the lowest minADE$_{1s}$ and minFDE$_{1s}$, but its mAP is relatively low, indicating weaker overall accuracy on multi-modal prediction compared to multi-horizon settings $\left\{I,S,M,L\right\}$. Combining all horizon-specific heads with the context head provides a balanced trade-off between short-term accuracy and long-term consistency, yielding more stable overall performance.

\begin{table}[!t]
\centering
\caption{{Ablation study on data expansions. $\mathcal{D}_S$ denotes the set from sensor augmentation, and $\mathcal{D}_C$ denotes the set from corrective behavior expansion. } 
}
\vspace{-0.5em}
\label{tab:data abl}
\resizebox{\linewidth}{!}{%
\begin{tabular}{ll ccccc}
\toprule
$\mathcal{D}_S$ & $\mathcal{D}_C$  & mAP~$\uparrow$ & L2$_{1s}$~$\downarrow$ &  L2$_{2s}$~$\downarrow$ & L2$_{4s}$~$\downarrow$ &  L2$_{8s}$~$\downarrow$ \\
\midrule
& &0.660 &0.381 &0.789 &1.677 &4.139\\
$\checkmark$& &0.670  &0.358 &0.748 &1.617 &3.940\\
&$\checkmark$ &0.679 &0.355 &0.754 &1.621 &4.028\\ 
$\checkmark$ &$\checkmark$ &\textbf{0.695}  &\textbf{0.342} &\textbf{0.700} &\textbf{1.434} &\textbf{3.718}\\ 
\midrule
$\checkmark$& &0.374 &0.914 &1.949 &4.817 &9.137\\
$\checkmark$ &$\checkmark$ &\textbf{0.565} &\textbf{0.645} &\textbf{1.348} &\textbf{4.499} &\textbf{8.562} \\ 
\bottomrule
\end{tabular}%
\vspace{-3mm}
}
\end{table}

\noindent\textbf{Effect of data expansions.} We conduct ablation studies on the effectiveness of different data expansion strategies. 
As shown in Tab.~\ref{tab:data abl}, each expansion individually improves performance over the baseline on SideWalks-Regular, and combining both yields the best results across all metrics, demonstrating their complementary benefits. Furthermore, on Sidewalks-Recovery, incorporating $\mathcal{D}_C$ significantly reduces both short-horizon and long-horizon errors, indicating that corrective behavior expansion enables the policy to learn from near-failure cases and recover from deviations.


\section{Real-World Deployment}
\label{sec:Real-World Deployment}



In this section, we present details of our real-world deployments with the wheeled robot\footnote{A demo video is available on the project page.}.

\subsection{Experimental Setup} We validate the effectiveness of the proposed approach across four environments, evaluated in both daytime and nighttime settings. The routes span different lengths (20m, 20m, 50m and 400m) to validate both short-horizon and long-horizon navigation performance. In each environment, a pedestrian walks across the path of the robot twice along the route to evaluate its performance in real-world sidewalk scenarios. For short-horizon trials, goal points are defined relative to the robot, while in long-horizon trials, GPS-based waypoints are used for continuous navigation. We use the success rates for goal reaching and pedestrian avoidance, and the success weighted by path length (SPL), for evaluation across all scenarios. In long-horizon navigation, we do not terminate the task when the robot goes off-route or collides. Instead, a human operator intervenes to take control, and we report the number of interventions as an additional metric in the 400m navigation task.

\begin{figure}[t!]
\centering
\includegraphics[width=\linewidth]{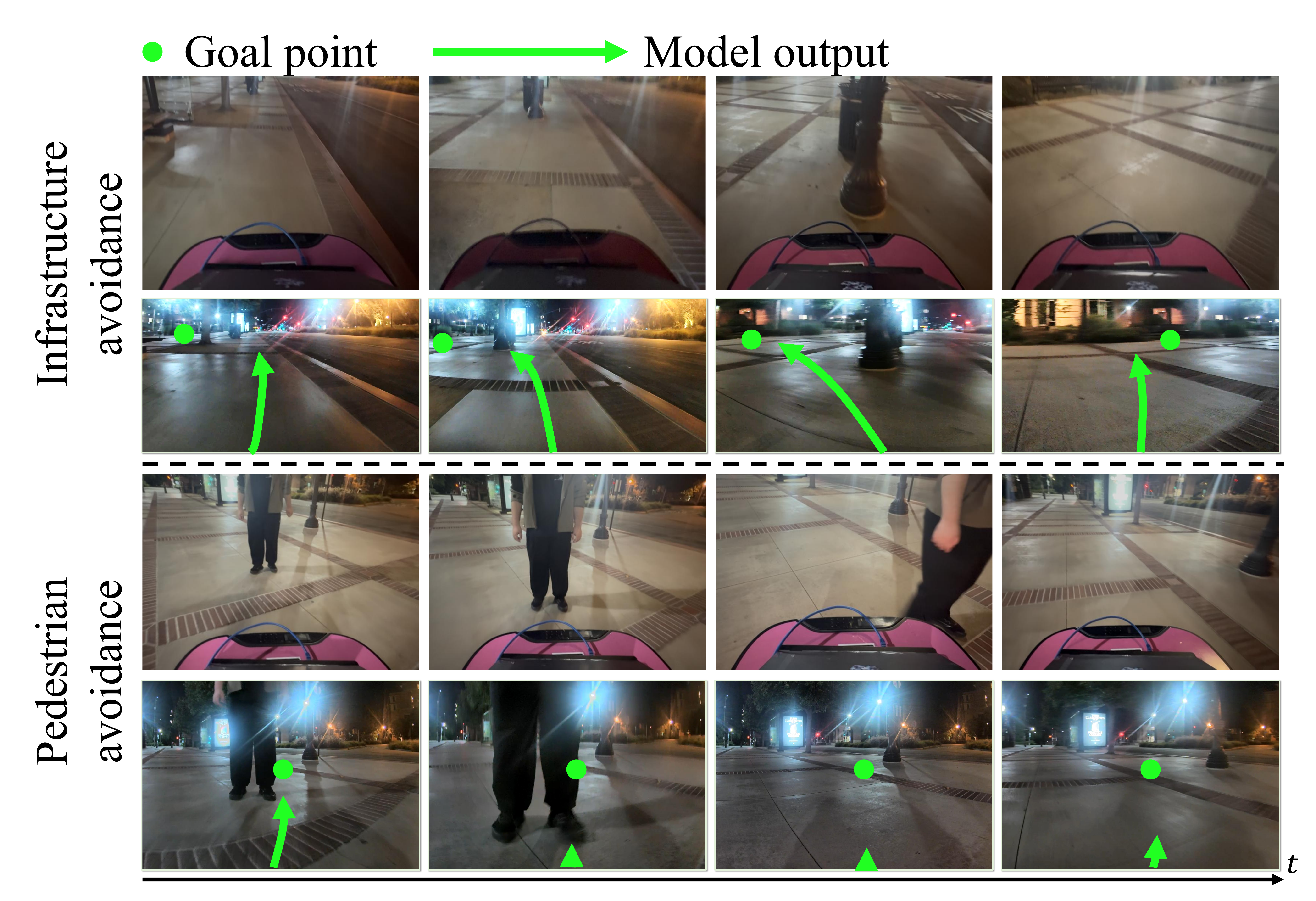}
\caption{
Qualitative results of MIMIC in the real-world.
}
\vspace{-3mm}
\label{fig:real-world}
\end{figure}

\begin{table}[t!]
\centering
\caption{{Closed-loop evaluation in the real world.} 
}
\vspace{-0.5em}
\label{tab:closedloop}
\resizebox{\linewidth}{!}{%
\begin{tabular}{l cc c|  c}
\toprule
 & \shortstack{Goal \\Reaching $\uparrow$} & \shortstack{Pedestrian \\Avoidance $\uparrow$} &\shortstack{SPL$\uparrow$ } & \shortstack{Intervention \\Times $\downarrow$} \\
\midrule
CityWalker &0.55 &0.18  &0.46&19  \\
CityWalker* &0.70 &0.29  &0.44&11  \\
\midrule
\ours MIMIC &\textbf{0.90} &\textbf{0.76} &\textbf{0.69}&\textbf{4}  \\ 
\bottomrule
\end{tabular}%
\vspace{-3mm}
}
\end{table}

\subsection{Results}

As illustrated in Tab.~\ref{tab:closedloop}, MIMIC outperforms CityWalker and its fine-tuned variant. MIMIC achieves the highest success rate in all navigation tasks. It requires far fewer intervention times, demonstrating the effectiveness of the proposed approach. We further provide qualitative results of two scenarios in Fig.~\ref{fig:real-world}. In the first scenario, the policy successfully navigates toward a goal point defined behind a tree: the robot turns to reach the target once sufficient space is available. In the second scenario, when a pedestrian is in front of the robot, the robot yields to avoid a collision.

\section{Conclusions and Future Work}
\label{sec:conclusion}

In this work, we present an imitation learning framework, MIMIC, for learning a sidewalk autopilot from the teleoperation dataset. First, we introduce corrective behavior expansion to extend the training distribution. Second, we propose using multi-scale, horizon-specific anchors for learning. We validate the proposed method on both the offline test set and real-world deployments, demonstrating its effectiveness.

\noindent\textbf{Limitations.} While MIMIC demonstrates its effectiveness, it also has limitations. Without explicit 3D or semantic supervision, the policy may degrade in highly cluttered or visually ambiguous environments. Introducing additional visual supervision, or distilling such knowledge from pretrained models, would be a promising direction.

\section{Acknowledgment}
\label{sec:ack}

The project was supported by the NSF Grants CNS-2235012 and IIS-2339769. Honglin He is supported by the Amazon Trainium Fellowship. We thank Coco Robotics for the generous donation of data and equipment.
{
\AtNextBibliography{\small}
\printbibliography
}

\end{document}